# Automatic detection of moving objects in video surveillance

Hanane BELHANI  
LaSTIC laboratory  
University of Batna 2  
Batna, Algeria  
hanane.belhani05@gmail.com

Larbi GUEZOULI  
LaSTIC laboratory  
University of Batna 2  
Batna, Algeria  
larbi.guezouli@univ-batna2.dz

*Abstract*—This work is in the field of video surveillance including motion detection. The video surveillance is one of essential techniques for automatic video analysis to extract crucial information or relevant scenes in video surveillance systems. The aim of our work is to propose solutions for the automatic detection of moving objects in real time with a surveillance camera. The detected objects are objects that have some geometric shape (circle, ellipse, square, and rectangle).

*Keywords-Video surveillance; motion detection; real-time system; pattern recognition.*

## I. INTRODUCTION

So far, CCTV systems (Closed Circuit TeleVision) become usable in several areas of security, among other artificial intelligence. In this regard, we will offer a video surveillance system that detects moving objects with some geometric form in real time.

Before starting to present our work, we will talk about problems of CCTV systems.

The detection of the background is the first step in motion detection process. Many conditions cause poor detection of the background as the lighting changes, repetitive movements, clutter and non-rigid objects (moving) appeared in the foreground. This poor background detection influences the motion detection of objects in the video especially in real-time detection. Therefore, the detection of a good background image facilitates the detection of moving objects reliably [1-5].

Several challenges may arise from the nature of video surveillance systems. These challenges are as follows [6]:

- Illumination changes (brightness): A basic model must be adapted to gradual and sudden changes in the appearance of the environment as well as the progressive illumination changes including the change in light intensity outdoors during the day (e.g., clouds moving).
- Dynamic background: A natural scene usually consists of dynamic objects. These dynamic objects can be composed by shaking trees, swaying curtains, undulating surface waters, waving flags, etc.
- Moving object: If a foreground object leaves the scene, it will create a ghost (regions that are detected as moving but do not correspond to moving objects), then the background model must adapt this object as back-plan. For example: If a parked car leaving the scene, the corresponding region should be accepted as part of the background.
- Video noise: The images video may contain brightness or color variations in video sequences called noises. A model of background extraction must face these degraded videos affected by different types of noises, such as sensor noise or compression artifacts.

Existing approaches have several limitations, which motivate our proposals [7-10]. Among the problems:

- The real-time operation.
- The detection of moving objects with some geometric shape (square, rectangle, circle).

For this, the main goal of our research is to provide solutions that detect moving objects with certain geometry in real time.

In our work, we seek to propose a method which can reduce the effect of the challenges described above and can run a fairly effective and efficient manner in terms of results and response time.

## II. PRESENTATION OF PROPOSED APPROACHES

We will give a presentation or a detailed description of the proposed approaches that detect the movement of objects with certain geometry in real time. We will present two different approaches.

### A. First approach

In this approach, we seek to detect non-rigid objects (moving) that have some geometric shape. Common fundamental steps to ensure movement detection process are:

- The first step is the detection of background, i.e. the image of the scene without movement (constituting stationary objects). Images acquired first time can be used to detect the model of the ideal background. The detection of the background is determined in an iterative manner to solve occlusions problems of objects in the image built by the foreground. The modified moving average [11] (MMA) is used to calculate the average of frames 1 to K for the generation of the model of the original background. For each pixel *(x,y)*, the corresponding value of the current background model $B_t(x,y)$ is calculated using the following formula:

$$B(x,y) = B_{t-1}(x,y) + 1/t \cdot (I_t(x,y) - B_{t-1}(x,y)) \quad (1)$$

Such as:

$B_{t-1}(x, y)$ is the previous background model.

$I_t(x, y)$ is the current video image captured.

*t* is the number of the captured image.

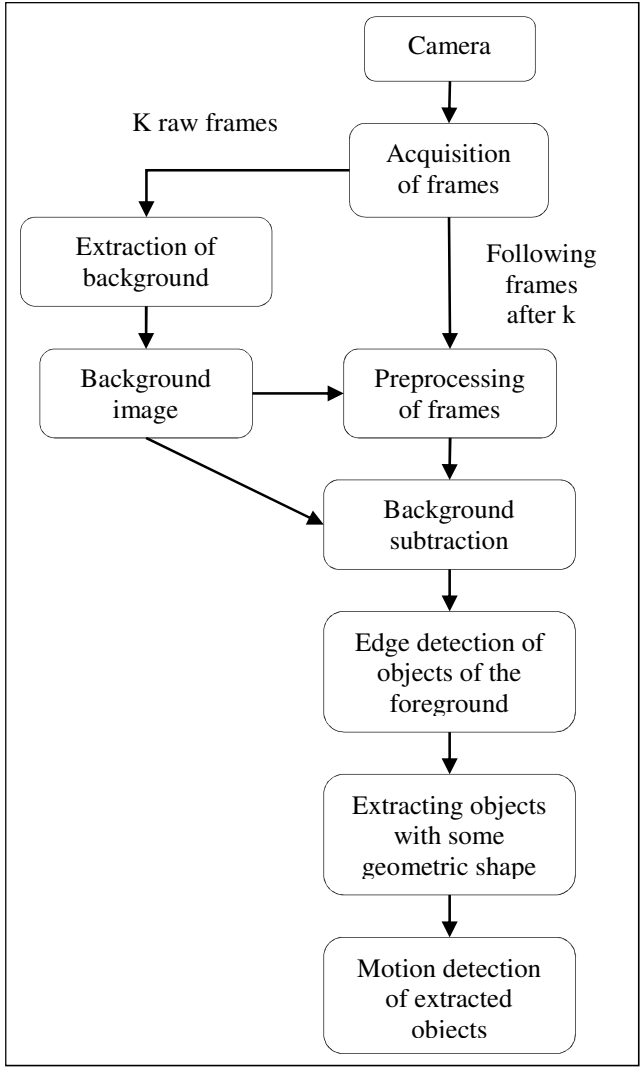

Figure 1. General scheme of first proposed approach.

- The second step is the processing of background image and images acquired after that. It is necessary to eliminate noises:
    - Color Space Conversion as GrayScale, YCrCb. These color spaces are expected to be more robust to shadows and brightness changes as RGB.
    - Application of smoothing (filtering or blurring) to eliminate noise.

- The third step is the subtraction of the background from the foreground, i.e. model the image of moving objects without background. Several algorithms are available [12]. This step requires the background and acquired image at time *t*. An absolute difference between the two images is applied to calculate the difference image. After, the resulting image is segmented by performing segmentation by thresholding:
    - If the pixel of the difference image is a pixel of the background, then it is replaced by a black pixel (0).
    - If the pixel of the difference image is a pixel of the foreground, then it is replaced by a white pixel (255).

$$d(x,y) = \begin{cases} 0 & |I_t(x,y) - B(x,y)| < \Delta t \\ 255 & |I_t(x,y) - B(x,y)| \geq \Delta t \end{cases} \quad (2)$$

- The threshold $\Delta t$ is used to determine if the pixel is a pixel of the background or of the foreground. It is obtained through a mathematical formula [11] as follows:

$$\Delta t = \lambda \cdot (1/N \cdot M) \cdot \sum_{x=1}^{N} \sum_{y=1}^{M} |I_t(x,y) - B(x,y)| \quad (3)$$

such as:

$\lambda$ is the coefficient of inhibition, where the value varies depending on the environment and the reference value is equal to 2, for this case, it is equal to 1.

$N \cdot M$ is the size of images.

$I_t$ is the current captured image.

$B$ is the image of the background created by the previous step.

This threshold reflects global changes in the scene, he takes a low value if there is a small change in brightness in the image and it increases if there are many changes.

Another preprocessing to improve the detection of objects in the foreground is needed. Unfortunately, some residual noise remains on the segmented difference image. We apply some morphological operations like dilation, erosion, opening and closing [13].

- The fourth step is the core of our work. It consists in the detection of moving objects that have some geometric shape i.e. based on geometric and spatial characteristics of objects. The following processes are applied:
    - The edge detection using Canny detector [14] which guarantees good detection (low error rate), good location and clarity of response (no false positives).
    - Information geometry (convex contours, the number of vertices or corners of the contours, the surface contours) are used to extract the contours of objects with some geometric shape (Circle, ellipse, square, rectangle).
    - The image of the detected objects is built. This image contains only extracted objects.
- The last step is the motion detection. Using one of the similarity measures (SAD: Sum of Absolute Differences)

between built images of detected objects (image at time *t* and image at time *t-1*).

In this step we detect the motion by using the approach of the sum of absolute differences (SAD) [15]. This approach is mainly used for measuring the similarity between two images. SAD is given by the following equation:

$$d(I_{t-1}, I_t) = \left(1/{N \cdot M}\right) \cdot \sum_{x=1}^{N} \sum_{y=1}^{M} |I_{t-1}(x,y) - I_t(x,y)| \quad (4)$$

such as:

$I_{t-1}$ is the built image at time (t-1).

$I_t$ is the built image at time (t).

$N \cdot M$ is the size of images.

A value of distance greater than the threshold means that there is a motion in the current image.

### B. Second approach

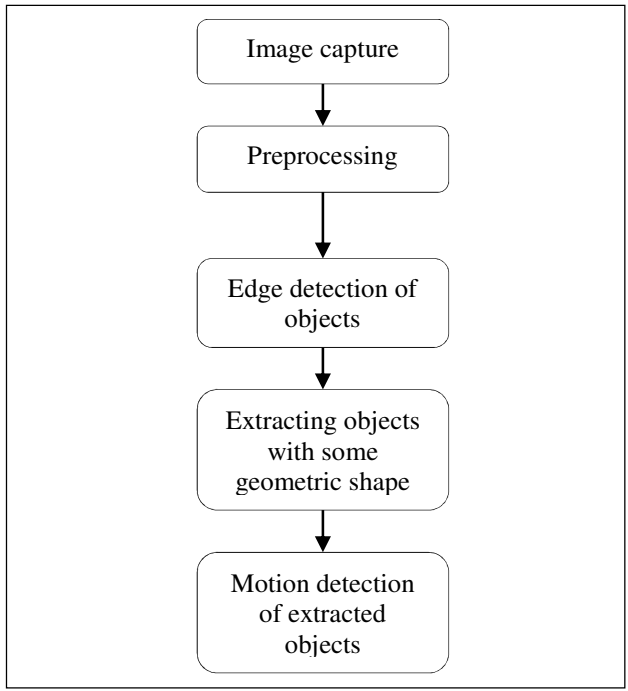

Figure 2. General scheme of the second proposed approach.

In this approach, we seek to detect non-rigid objects (moving) that have some geometric shape by using captured images.

The process of motion detection is as follows:

- The first step is the preprocessing of acquired images that is necessary to remove noise and other problems:
  - Color Space Conversion as GrayScale, YCrCb. These color spaces are expected to be more robust to shadows and brightness changes as RGB.
  - Application of smoothing to eliminate noise.
- The second step is the detection of outlines using Canny detector [14]. We need only convex and closed contours.
- The third step is the extraction of objects that have some geometric shape (square, rectangle, circle, and ellipse).

The extraction of objects is based on geometric characteristics of contours (the convex contours, the number of contour vertices, calculation of angles, and the surface contour) for sensing shapes of objects. The image of detected objects is built. This image contains extracted objects.

- The last step is the motion detection process using constructed images (image at time t and image at time t-1) by the previous step. Motion detection is done as in the first approach.

### III. IMPLEMENTATION EVALAUTION AND EXPERIMENTS

#### A. Work tools

The used programming language is C++ with Microsoft Visual Studio 2013 programming environment and OpenCV library 2.4.10 (Open Source Computer Vision Library).

#### B. Evaluation and experiments

*1) Evaluation tools and tests*

To evaluate our work we used several software and hardware tools as shown in Table I.

TABLE I. SOFTWARE AND HARDWARE EVALUATION TOOLS.

| Software Tools | |
|---|---|
| Operating system | Microsoft Windows 8.1 Professional |
| Compiler | C++ under Microsoft Visual Studio Ultimate 2013 |
| Library | OpenCv 2.4.10 |
| **Hardware Tools** | |
| Processor | Intel(R) Core(TM) i3-380M (2.53 GHz) |
| RAM | 6.00 Go |
| Camera | 1.3M HD Webcam |

*2) Results*

We present in this section results obtained by different experiments applied to our system.

Figure 3 shows the implementation of the first approach for detection and subtraction of the background. The goal is to detect moving objects with some geometric shape (square, rectangle, circle, and ellipse).

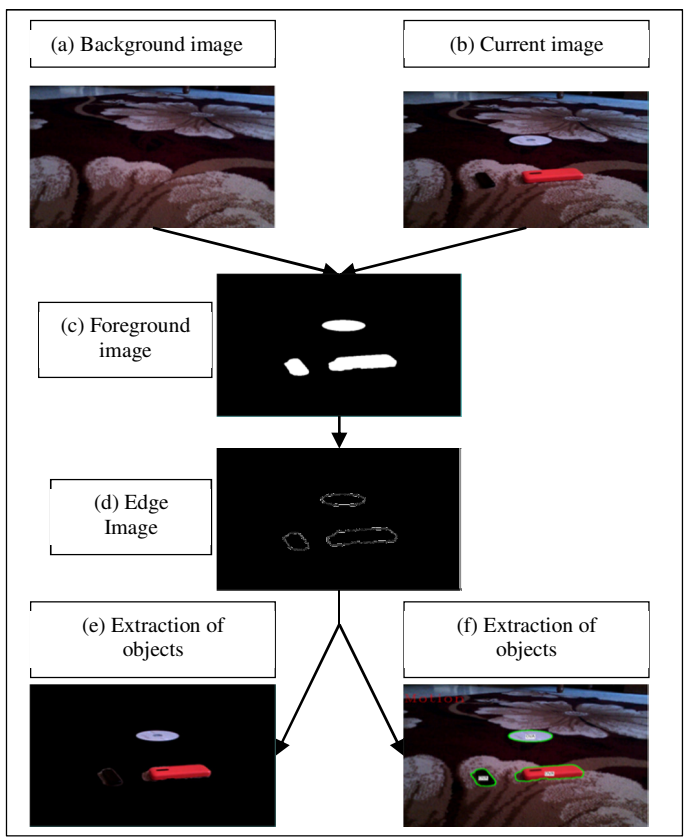

Figure 3. Scheme of the first approach.

In Figure 4, we present the implementation of the second approach using captured images to detect edges used in extracting process of moving objects with some geometric shape.

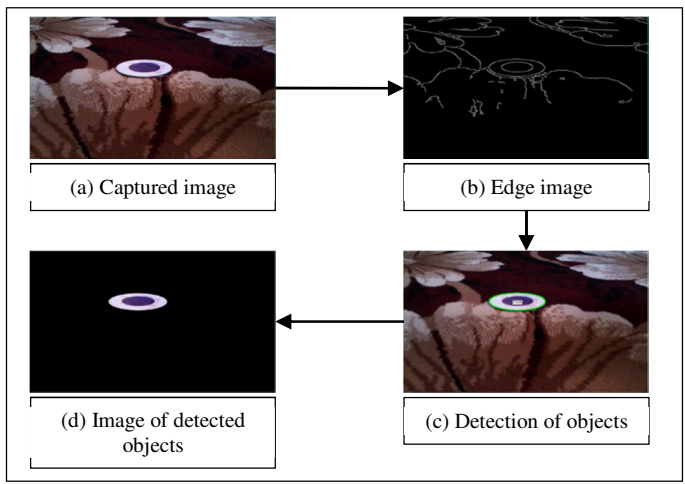

Figure 4. Scheme of the second approach.

*3) Evaluation*

To check performances of our approaches, we did some experiments and assessments.

   *a) Execution time and complexity*

The execution time is a very importance in our system because we need a real-time system. We present here the average of execution time (in sequential and parallel cases) of our system.

TABLE 2. THE TIME MEAN OF BASIC SYSTEM MODULES.

| Modules | Sequential time | Parallel time |
|---|---|---|
| BackgroundDetector | 9.89 sec | 3.36 sec |
| ForegroundDetector | 0.62 sec | 0.15 sec |
| MotionDetector | 0.70 sec | 0.14 sec |

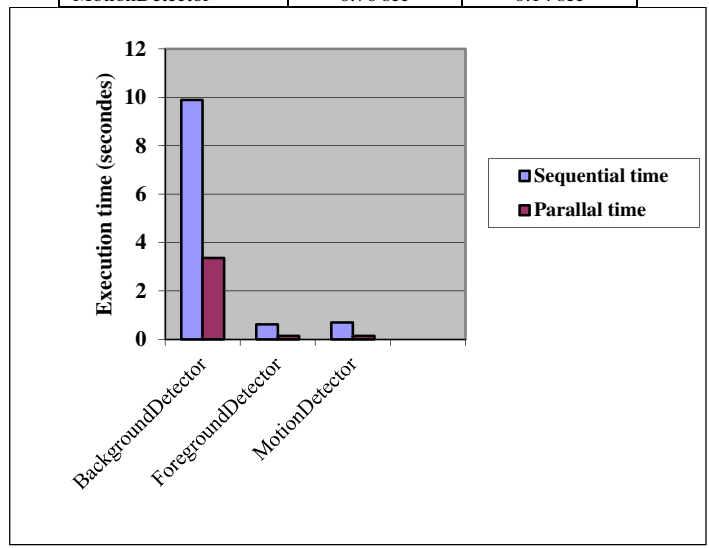

Figure 5. Comparison between sequential and parallel execution time.

   *b) Detection time*

The detection time of a moving object is dependent upon detection of its contour. This detection depends on other characteristics inter alia the speed of the object. The speed has great influence on the appearance of the object in a clearer way to detect its outline and shape. It can be calculated based on the position of the object at the two moments: previous and current time. The position of the object is extracted using spatial moments (HuMoments) [16].

$$V = \frac{\Delta p}{\Delta t} = \frac{(P_i - P_{i-1})}{(T_i - T_{i-1})} \qquad (5)$$

such as:

*V*: The speed of the object.

$P_{i-1}$ and $P_i$ are the positions of the center of object in two captured images where it appeared (the current image *i* where it is detected, and the previous *i-1*).

$T_i$ and $T_{i-1}$: Times of images *i* and *i-1*.

In our experiments, we try to calculate the speed limit where the system can detect the edge of an object and its form. We use an example of a video sequence captured with an object in circular motion. Table 3 and Table 4 present the values of the distance between two successive positions of the object, the difference in time and speed. The distance is calculated is in pixels, then it is converted to meters (m) using drawing scale. The time is presented in seconds (sec) and the speed in (m / sec).

TABLE 3. VALUES OF DISTANCE TRAVELED, TIME CONSUMED AND SPEED OF A MOVING IN A VIDEO SEQUENCE (FIRST APPROACH).

| $p_i$-$p_{i-1}$ | 0.2 | 0.48 | 0.36 | 0.68 | 0.50 | 0.80 | 0.96 | 0.8 | No detection |
|---|---|---|---|---|---|---|---|---|---|
| $t_i$-$t_{i-1}$ | 0.5 | 1.01 | 0.6 | 0.99 | 0.52 | 0.62 | 0.66 | 0.53 | |
| speed | 0.4 | 0.47 | 0.6 | 0.69 | 0.96 | 1.29 | 1.45 | 1.5 | |

TABLE 4. VALUES OF DISTANCE TRAVELED, TIME CONSUMED AND SPEED OF A MOVING OBJECT IN A VIDEO SEQUENCE (SECOND APPROACH).

| $p_i$-$p_{i-1}$ | 0.004 | 0.02 | 0.025 | 0.36 | 0.06 | 0.6 | 0.68 | 0.8 | No detection |
|---|---|---|---|---|---|---|---|---|---|
| $t_i$-$t_{i-1}$ | 0.23 | 0.23 | 0.22 | 1.52 | 0.23 | 0.89 | 0.89 | 0.67 | |
| speed | 0.02 | 0.09 | 0.11 | 0.24 | 0.27 | 0.67 | 0.76 | 1.19 | |

According to these results, we found that the first approach cannot detect moving objects with speed greater than 1.5 m/sec and the second one cannot detect moving objects with speed greater than 1.19 m/sec.

*c) Metric evaluation*

We used recall *r* and precision *p* measures and F1 measure to evaluate the performance of our system.

The definitions of these measures and their formulas are given as follows:

- **Recall**: The ratio between the number of correctly detected objects which have the desired geometric shape and the total number of existing objects in the scene which have the desired geometric shape.

$$Recall = \frac{TP}{TP+FN} \qquad (6)$$

- **Precision**: The ratio between the number of correctly detected objects which have desired geometric shape and the total number of detected objects (which have or have not the desired geometric shape).

$$Precision = \frac{TP}{TP+FP} \qquad (7)$$

such as:

*TP*: The number of correctly detected objects which have desired geometric shape.

*FN*: The number of objects in the scene that are not detected and have the desired geometric shape.

*FP*: The total number of detected objects of the scene which don't have the desired geometric shape.

- **F1-measure**: The score can be interpreted as a weighted average of the precision and recall.

$$F_1 = 2 \times \frac{precision \cdot recall}{precision+recall} \qquad (8)$$

$F_1$ reaches its best value to 1 and its bad value to 0.

To calculate both measures recall and precision, we use video as images captured in different rooms where there are objects with different positions (close, far) to the camera and different forms (small, large). Thus, different features are used: color of objects (light or dark), number of objects, background (one or more colors, light or dark) and speed of objects).

We present results of different tests carried out in the following tables. We use video sequences where each sequence is a set of captured images.

Table 5 and Table 6 present values of different measures (recall, precision and F1-measure) obtained from the two proposed approaches.

TABLE 5. VALUES OF THE RECALL, PRECISION AND F1-MEASURE OBTAINED WITH THE FIRST APPROACH.

| Scenes | Number of appearances of sought objects | TP | FP | FN | R | P | F1 |
|---|---|---|---|---|---|---|---|
| One object in the scene | 150 | 124 | 1 | 26 | 0.83 | 0.99 | 0.90 |
| Moving objects | 287 | 271 | 28 | 16 | 0.94 | 0.91 | 0.92 |
| Object with some sides | 150 | 115 | 20 | 35 | 0.77 | 0.85 | 0.81 |
| Brightness changing | 150 | 112 | 30 | 38 | 0.75 | 0.79 | 0.77 |

By examining results of the Table 5, we note that:

- Obtained values show that the first proposed approach based on the subtraction of the background allows the detection of objects with some geometric shape (circular or quadrilateral).
- Despite the *FP* and *FN*, *F1* is better and close to 1.
- The appearance of false positives is due to the occlusion of other objects, shadows or in the case of brightness change.

TABLE 6. VALUES OF THE RECALL, PRECISION AND F1-MEASURE OBTAINED WITH THE SECOND APPROACH.

| Scenes | Number of appearances of objects sought | TP | FP | FN | R | P | F1 |
|---|---|---|---|---|---|---|---|
| One moving object | 150 | 120 | 9 | 30 | 0.80 | 0.93 | 0.86 |
| Objects within other objects | 350 | 290 | 20 | 60 | 0.83 | 0.93 | 0.88 |
| Lots of selected contours | 150 | 116 | 140 | 34 | 0.77 | 0.45 | 0.57 |
| Object with some sides (e.g. cube) | 150 | 60 | 35 | 90 | 0.40 | 0.63 | 0.49 |

By examining results of the Table 6, we note that:

- Obtained values show that the second proposed approach based on captured images allows the detection of objects with some geometric shape (circular or quadrilateral). Its effect is especially apparent in the detection of objects that are within other objects.

- For precision values in the two first cases, they are better because of the detection of false positives FP is minimal (not many false contours).

- There are a lot of detected contours. The measurement values (recall, precision and F1 score) are low because of the appearance of FP.

- Also, for items that have some sides, obtained measurement values are low, and it shows that the detection of these objects can be decreased due to the non-detection of contours during their movement.

## IV. CONCLUSION

This work deals with the field of video surveillance systems. These systems can be used in many application areas: security of the premises, detection of accidents, fires, robotics, object recognition ... The video is the media treated in such systems. Among the most important steps in video surveillance systems the motion detection. This step involves the detection of moving objects in video sequences captured by the surveillance camera. The motion detection stage is among the most studied problems in the field of video analysis where many research works focus on this problem.

Our work focuses on the problem of detecting moving objects with some geometric shape. This detection is done in real time. As part of this work, we have proposed two approaches:

- The first approach focuses on two basic steps: modeling the background to build the image of the scene (without moving objects) and subtracting the background from the foreground image which allows getting moving objects.

- The second approach exerts directly on captured images. The edge detection step is applied to extract only edges of objects having required forms (circular or quadrilateral).

- Each of the two approaches has some advantages and limitations.

- With the aim of making our system as a real-time one, we applied the parallelism to some modules of the system (in both proposed approaches).

A set of experiments was conducted to assess the performances of proposed approaches in terms of detection of objects with popular shapes (the limit of the speed of objects, recall and precision). Our evaluations have shown that proposed approaches were able to detect moving objects with sought forms despite noticed limitations.